# A Novel Statistical Fusion Rule for Image Fusion and its Comparison in Non Subsampled Contourlet Transform Domain and Wavelet Domain

Manu V T [1] and Philomina Simon [2]

[1]Department of Computer Science, University of Kerala, Kariavattom Campus, Thiruvananthapuram, 695 581, Kerala, India
manuvt.nitc@gmail.com

[2]Department of Computer Science, University of Kerala, Kariavattom Campus, Thiruvananthapuram, 695 581, Kerala, India
philomina.simon@gmail.com

## ABSTRACT

Image fusion produces a single fused image from a set of input images. A new method for image fusion is proposed based on Weighted Average Merging Method (WAMM) in the Non Subsampled Contourlet Transform (NSCT) domain. A performance analysis on various statistical fusion rules are also analysed both in NSCT and Wavelet domain. Analysis has been made on medical images, remote sensing images and multi focus images. Experimental results shows that the proposed method, WAMM obtained better results in NSCT domain than the wavelet domain as it preserves more edges and keeps the visual quality intact in the fused image.

## 1. INTRODUCTION

Image fusion provides an efficient way to merge the visual information from different images. The fused image contains complete information for better human or machine perception and computer-processing tasks, such as segmentation, feature extraction, and object recognition. Image fusion can be done in pixel level, signal level and feature based. The traditional image fusion schemes performed the fusion right on the source images, which often have serious side effects such as reducing the contrast. Later researchers realized the necessity to perform the fusion in the transform domain as mathematical transformations provides further information from the signal that is not readily available in the raw signal.

With the advent of wavelet theory, the concept of wavelet multi-scale decomposition is used in image fusion [9]. The wavelet transform has been used in many image processing applications such as restoration, noise removal, image edge enhancement and feature extraction; wavelets are not very efficient in capturing the two-dimensional data found in images[5]. Several transform have been proposed for image signals that have incorporated directionality and multiresolution and hence, those methods could not efficiently capture edges in natural images. Do and Vetterli proposed contourlet transform[8], an efficient directional multi resolution image representation. The contourlet transform achieves better results than discrete wavelet transform in image processing in geometric transformations. The contourlet transform is shift-variant based on sampling. However, shift invariance is a necessary condition in image processing applications.

The NSCT is a fully shift-invariant, multiscale and multidirection expansion that has a fast implementation[1]. It achieves a similar sub band decomposition as that of contourlets, but without downsamplers and upsamplers in it, thus overcoming the problem of shift variance[2].

## 2. NON SUBSAMPLED CONTOURLET TRANSFORM

The Non Subsampled Contourlet Transform (NSCT) is constructed by combining the Non subsampled Pyramids (NSP) and the Non subsampled Directional Filter Banks (NSDFB)[1]. The former provide multiscale decomposition and the later provide directional decomposition[3]. A Non subsampled Pyramid split the input into a low-pass subband and a high-pass subbands. Then a Non subsampled Directional Filter Banks decomposes the high-pass subband into several directional subbands. The scheme is iterated repeatedly on the low-pass subband [11].

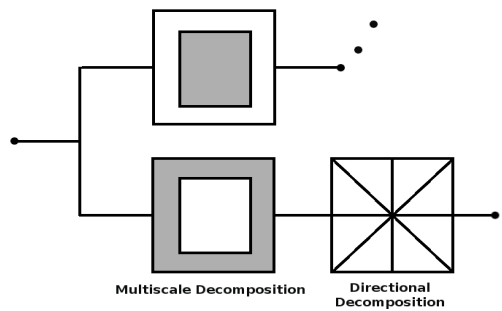

Figure 1. Block Diagram of NSCT

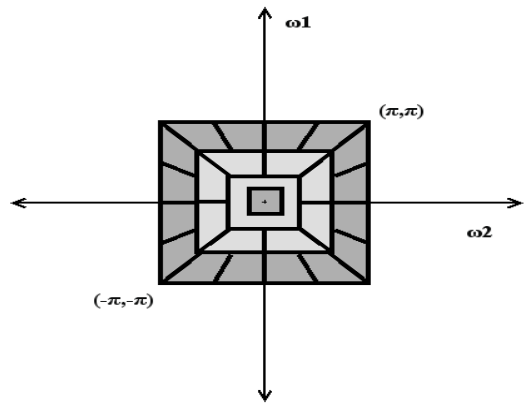

Figure 2. Block Diagram Frequency division

## 3. IMAGE FUSION SCHEME AND STATISTICAL FUSION RULES

Image fusion scheme in two source images can be considered as a step by step process. First, the source images are divided into coarse scales and fine scales. Coarse scales represent the high frequency components and fine scales represent low frequency components in the source images. Low frequency components contain overall details of the image while the high frequency components contain details about edges and textures. Then, the coefficients of the source images

The International Journal of Multimedia & Its Applications (IJMA) Vol.4, No.2, April 2012are decomposed. Second, the coarse scales and the fine scales in the source images are separately fused based on statistical fusion rule using NSCT [4][5][6][7]. Separate fusion rules are applied on these fine scales and coarse scales to obtain the fusion coefficients. The fused image is obtained by inverse NSCT from these fusion coefficients.

In this section, two different statistical fusion rules are discussed. These rules are analyzed experimentally thereby examining the performance of the image fusion in both wavelet and NSCT domain.

### 3.1. Method 1- Fusion based on Entropy

Entropy is the measure of information content in the image. A high value of entropy denotes more information content and vice versa. So this statistical measure could be used in making a decision to select the fusion coefficient.

$$H(S) = -\sum P(X) \log P(X) \qquad 1$$

Entropy is calculated on the low frequency components of the input images within a 3-by-3 window and whichever having higher values of entropy were selected as the fusion coefficients among the low frequency components

For the high frequency components, regional energy is calculated over a 5-by-5 window using the formula

$$E_k(i,j) = \sum_{m=-2}^{2} \sum_{n=-2}^{2} \sum_{s=-2}^{S} \sum_{d=-2}^{D} W(m+3, n+3) \bullet (C_K^{\{s,d\}}(i+m, j+n))^2 \qquad 2$$

where $C_K^{\{s,d\}}$ is the NSCT coefficient corresponding to scale *s* and direction *d* at position *(i,j)* for the image *k*.

W is a filter that gives more weightage to the central coefficient and is defined as

$$W = \frac{1}{256} \begin{bmatrix} 4 & 4 & 4 & 4 & 4 \\ 4 & 16 & 16 & 16 & 4 \\ 4 & 16 & 64 & 16 & 4 \\ 4 & 16 & 16 & 16 & 4 \\ 4 & 4 & 4 & 4 & 4 \end{bmatrix} \qquad 3$$

Then the coefficient is chosen as the fuse coefficient when the region energy of it is larger shown as formula

$$C_F^{\{s,d\}}(i,j) = \begin{cases} C_A^{\{s,d\}}(i,j) & E_A(i,j) \geq E_B(i,j) \\ C_B^{\{s,d\}}(i,j) & otherwise \end{cases} \qquad 4$$

71

Finally the fused image is reconstructed using the fused coefficients, $C_F^{\{s,d\}}$ using the inverse NSCT transform.

### 3.2. Method 2- Fusion based on Mean

Mean is the representative value of a large dataset that describes the center or middle value. Mean is the measure of the group contributions per contributor which is conceived to be the same as the amount contributed by each *n* contributors if each were to contribute equal amounts

$$\bar{x} = \frac{1}{n}\sum_{i=1}^{n} x_i \qquad 5$$

Mean is calculated on the low frequency components of the input images within a 3-by-3 window and whichever having higher values of mean were selected as the fusion coefficients among the low frequency components.

For the high frequency components, regional energy is calculated over a 5-by-5 window using the formula

$$E_k(i,j) = \sum_{m=-2}^{2}\sum_{n=-2}^{2}\sum_{s=-2}^{S}\sum_{d=-2}^{D} W(m+3, n+3) \bullet (C_K^{\{s,d\}}(i+m, j+n))^2 \qquad 6$$

where $C_K^{\{s,d\}}$ is the NSCT coefficient corresponding to scale *s* and direction *d* at position *(i,j)* for the image *k*.

W is a filter that gives more weightage to the central coefficient and is defined as

$$W = \frac{1}{256}\begin{bmatrix} 4 & 4 & 4 & 4 & 4 \\ 4 & 16 & 16 & 16 & 4 \\ 4 & 16 & 64 & 16 & 4 \\ 4 & 16 & 16 & 16 & 4 \\ 4 & 4 & 4 & 4 & 4 \end{bmatrix} \qquad 7$$

Then the coefficient is chosen as the fuse coefficient when the region energy of it is larger shown as formula

$$C_F^{\{s,d\}}(i,j) = \begin{cases} C_A^{\{s,d\}}(i,j), & E_A(i,j) \geq E_B(i,j); \\ C_B^{\{s,d\}}(i,j), & otherwise \end{cases} \qquad 8$$

Finally the fused image is reconstructed using the fused coefficients, $C_F^{\{s,d\}}$ using the inverse NSCT transform.

### 3.3. Method 3- Fusion based on Standard Deviation

Standard Deviation provides a way to determine regions which are clear and vague. It is calculated by the formula

$$s = \sqrt{\frac{1}{n}\sum_{i=1}^{n}(x_i - \tilde{x})^2}$$
9

where

$$\tilde{x} = \frac{1}{n}\sum_{i=1}^{n} x_i$$
10

Standard Deviation is calculated on the low frequency components of the input images within a 3-by-3 window and whichever having higher values of mean were selected as the fusion coefficients among the low frequency components.

For the high frequency components, regional energy is calculated over a 5-by-5 window using the formula

$$E_k(i,j) = \sum_{m=-2}^{2}\sum_{n=-2}^{2}\sum_{s=-2}^{S}\sum_{d=-2}^{D} W(m+3,n+3) \bullet (C_K^{\{s,d\}}(i+m, j+n))^2$$
11

where $C_K^{\{s,d\}}$ is the NSCT coefficient corresponding to scale *s* and direction *d* at position *(i,j)* for the image *k*.

W is a filter that gives more weightage to the central coefficient and is defined as

$$W = \frac{1}{256}\begin{bmatrix} 4 & 4 & 4 & 4 & 4 \\ 4 & 16 & 16 & 16 & 4 \\ 4 & 16 & 64 & 16 & 4 \\ 4 & 16 & 16 & 16 & 4 \\ 4 & 4 & 4 & 4 & 4 \end{bmatrix}$$
12

Then the coefficient is chosen as the fuse coefficient when the region energy of it is larger shown as formula

$$C_F^{\{s,d\}}(i,j) = \begin{cases} C_A^{\{s,d\}}(i,j) & E_A(i,j) \geq E_B(i,j) \\ C_B^{\{s,d\}}(i,j) & otherwise \end{cases}$$
13

Finally the fused image is reconstructed using the fused coefficients, $C_F^{\{s,d\}}$ using the inverse NSCT transform.

## 4. IMAGE FUSION BASED ON WEIGHTED AVERAGE MERGING METHOD (WAMM)–PROPOSED APPROACH

In this section, we discuss the fusion based on WAMM in NSCT Domain. WAMM is used in the high frequency components to obtain the fusion coefficient whereas Standard Deviation is calculated on the low frequency components of the input images within a 3-by-3 window. An average of the low frequency components is calculated. Whichever obtains the higher values of average are selected as the fusion coefficients among the low frequency components.

The main features of this new method are that it preserves the image quality and the edge details of the fused image. The visual quality of the fused image is better in NSCT domain.

The Weighted Average Merging Method (WAMM) is formulated as

$$\begin{cases} C_F^{\{s,d\}}(i,j) = w_{max} C_A^{\{s,d\}}(i,j) + w_{min} C_B^{\{s,d\}}(i,j), & E_A(i,j) \geq E_B(i,j); \\ C_F^{\{s,d\}}(i,j) = w_{min} C_A^{\{s,d\}}(i,j) + w_{max} C_B^{\{s,d\}}(i,j), & E_A(i,j) < E_B(i,j); \end{cases} \quad 14$$

The weights are estimated as:

$$\begin{cases} W_{min} = 0, W_{max} = 1, & M_{AB}(p) < T; \\ W_{min} = \frac{1}{2} - \frac{1}{2}\left(\frac{1-M_{AB}}{1-T}\right), W_{max} = 1 - W_{min}, & other \end{cases} \quad 15$$

Where T denote the threshold and $T \in (0, 0.5)$.

When the weight is zero, this means the substitution of an image by another.

$M_{AB}(p)$ is called the match measure which is defined as

$$M_{AB}(p) = \frac{2 \sum_{s \in S, t \in T} w(s,t) C_A^{\{s,d\}}(m+s,n+t) C_B^{\{s,d\}}(m+s,n+t)}{E_A(p) + E_B(p)} \quad 16$$

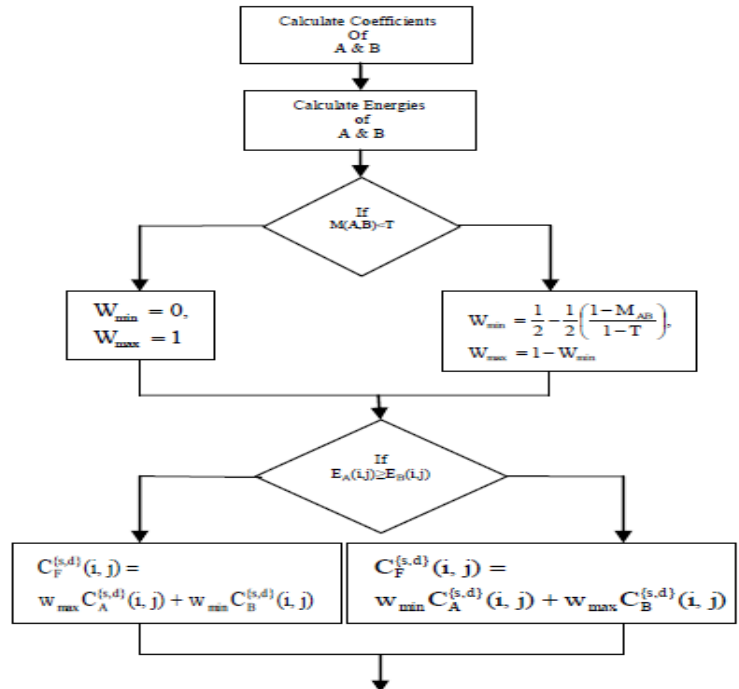

Figure 3: Schematic Diagram of WAMM

## 4.1 Fusion Based on Proposed Method in NSCT Domain

Standard Deviation is calculated on the low frequency components of the input images within a 3-by-3 window and whichever has the higher values of Mean are selected as the fusion coefficients among the low frequency components.

While WAMM is used in the high frequency components to obtain the fusion coefficient . The high frequency component has the crucial information within the images like the texture, brightness and contrast. The WAMM takes care of preserving these details much more better than the fusion schemes that we discussed before.

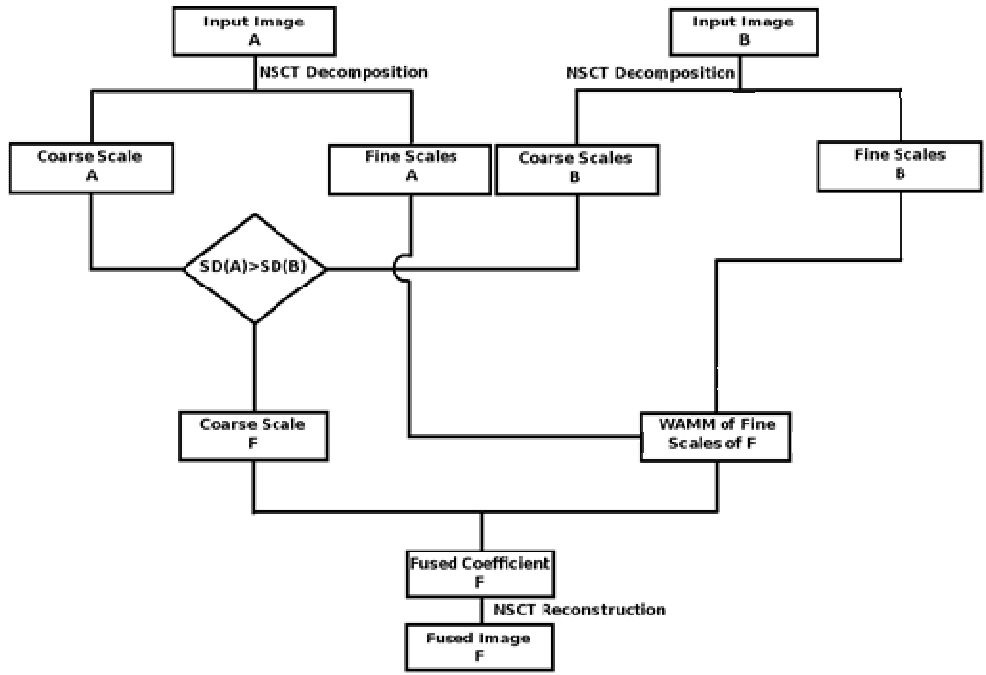

Figure 4: Schematic Diagram of Proposed Method

The use of standard deviation enhances the fusion scheme by preserving the edges in the images, while WAMM help to preserve the texture and other detailed information in the images by providing a way to combine the values without much bias, ie, a low value is compensated by giving a proper weight to provide a considerable contribution with respect to a high value.

## 5. PERFORMANCE MEASURES

The issue in the performance evaluation of an image fusion algorithm is that the unavailability of reference images. In addition, relevant research shows that a single measurement cannot be effectively evaluate the performance of different fusion algorithms or always cannot be consistent with human visual perception.

(1) Qualitative approaches: involves visual comparison of the input images and the output image.

(2) Quantitative approaches: involves a set of pre-defined quality indicators for measuring the spectral and spatial similarities between the fused image and the original images.

Because qualitative approaches and visual evaluations may contain subjective factor and may be influenced by personal preference, quantitative approaches are often required to prove the correctness of the visual evaluation.

For quantitative evaluation, a variety of fusion quality assessment methods have been introduced by different authors. The quality indexes/indicators introduced include, for example, Standard Deviation (SD), Mean Absolute Error (MAE), Root Mean Square Error (RMSE), Sum Squared Error (SSE) based Index, Agreement Coefficient based on Sum Squared Error (SSE), Mean Square Error (MSE) and Root Mean Square Error, Information Entropy, Spatial Distortion Index, Mean Bias Error (MBE), Bias Index, Correlation Coefficient (CC), Warping Degree (WD),Piella Metric (P), Spectral Distortion Index (SDI), Image Fusion Quality Index (IFQI), Spectral Angle Mapper (SAM), Relative Dimensionless Global Error (ERGAS),etc. However, it is also not easy for a quantitative method to provide convincing measurements.

In our work, objective analysis of the proposed method is done using the performance metrics. Even though these metrics do not provide a foolproof estimate of the performance of the method, they can be used in comparative analysis as a performance indicator. Following are the metrics used for performance evaluation in this research work.

### 5.1. Entropy

Entropy is a measure of information content of an image. It helps to know the information content of the source images and the output fused image. An increased value of entropy of the fused image implies a better fusion scheme.

$$H(S) = -\sum P(X) \log P(X) \qquad 17$$

### 5.2. Similarity Measure

The magnitude of gradient G(m, n) at a point (m, n) of image F is obtained by

$$G(m, n) = \frac{1}{2}\{|F(m,n) - F(m+1, n+1)| + |F(m,n) - F(m+1, n)|\} \qquad 18$$

where $G_1$, $G_2$ are the gradient images of input images. Then $G_1$, $G_2$ are combined into $G''$ by taking the maximum gradient value at each position. $G''$ can be seen as the gradient image of the ideal fusion image. The gradients of the actual fusion image $G$ are also calculated. The similarity S between the ideal fusion image and the actual fused image is calculated by formula

$$S(G, G') = 1 - \frac{\sqrt{\sum (G(m,n) - G'(m,n))^2}}{\sqrt{\sum (G(m,n))^2} + \sqrt{\sum (G'(m,n))^2}} \qquad 19$$

A higher value of S would indicate a better fusion scheme.

### 5.3. Piella Metric

Let $\mathbf{x}=\{x_i | i=1,2,\ldots,N\}$ and $\mathbf{y}=\{y_i | i=1,2,\ldots,N\}$ be the original and the test image signals, respectively. The Universal Image Quality index proposed by Zhou Wang [13] is defined as

76

The International Journal of Multimedia & Its Applications (IJMA) Vol.4, No.2, April 2012

$$Q = \frac{4\sigma_{xy}\bar{x}\bar{y}}{(\sigma_x^2 + \sigma_y^2)[(\bar{x})^2 + (\bar{y})^2]},$$

20

$$\bar{x} = \frac{1}{N}\sum_{i=1}^{n} x_i$$

21

$$\bar{y} = \frac{1}{N}\sum_{i=1}^{n} x_i$$

22

$$\sigma_x^2 = \frac{1}{N-1}\sum_{i=1}^{N}(x_i - \bar{x})^2$$

23

$$\sigma_y^2 = \frac{1}{N-1}\sum_{i=1}^{N}(y_i - \bar{y})^2$$

24

$$\sigma_{xy} = \frac{1}{N-1}\sum_{i=1}^{N}(x_i - \bar{x})(y_i - \bar{y})$$

25

This quality index models any distortion as a combination of three different factors: Loss of Correlation, Luminance Distortion, and Contrast Distortion. In order to understand this, the definition of Q can be rewritten as a product of three components:

$$Q = \frac{\sigma_{xy}}{\sigma_x \sigma_y} \cdot \frac{2\bar{x}\bar{y}}{(\bar{x})^2 + (\bar{y})^2} \cdot \frac{2\sigma_x \sigma_y}{\sigma_x^2 + \sigma_y^2}$$

26

The first component is the correlation coefficient between x and y, which measures the degree of linear correlation between x and y, and its dynamic range is [-1,1]. The best value 1 is obtained when $y_i = ax_i + b$ for all $i=1,2,...,N$, where $a$ and $b$ are constants and $a>0$. Even if x and y are linearly related, there still might be relative distortions between them, which are evaluated in second and third components. The second component, with a value range of [0,1], measures how close the mean luminance is between x and y. It equals 1 if and only if $\bar{x} = \bar{y}$. $\sigma_x$ and $\sigma_y$ can be viewed as estimate of the contrast of x and y, so the third component measures how similar the contrasts of the images are. Its range of values is also [0,1], where the best value 1 is achieved if and only if $\sigma_x = \sigma_y$.

Piella Metric[13] is a quality measure which is derived from the above mentioned metric and offers much more focus on the locality of reference of the images. It takes into account region-based measurements to estimate how well the important information in the source images is represented by the fused image. The evaluation of Piella Metric of the fused image *f* of the input images *a* and *b* is defined as

77

$$Q_E(a,b,f) = Q_W(a,b,f) \cdot Q_W(a',b',f')^\alpha \qquad 27$$

where $a', b', f'$ are the edge images of *a, b* and *f* respectively. Canny operator is selected to detect the edge information, which detects the edges by searching the local maximum of image gradient. Canny operator detects the strong edges and weak edges with two thresholds, respectively, where the thresholds are system automatic selection. The Canny operator is not sensitive to noise and can detect the true weak edges.

In order to measure the metric, a sliding window is employed: starting from the top-left corner of the two images, a sliding window of a fixed size traverses over the entire image until the bottom-right corner is reached. For each window the local quality index is computed. Finally, the overall image quality index is computed by averaging all local quality indices.

$Q_W(a,b,f)$ is the *weighted fusion quality index* and is defined as

$$Q_W(a,b,f) = \sum_{w \in W} c(w)\left(\lambda(w)Q_0(a,f|w) + (1-\lambda(w))Q_0(b,f|w)\right) \qquad 28$$

where $\lambda(w)$ is a local weight given by

$$\lambda(w) = \frac{s(a|w)}{s(a|w) + s(b|w)} \qquad 29$$

where $s(a|w)$ is some saliency of image *a* in window w. The energy is selected as the salient feature and the size of the window is 3 by 3 and it is moved starting from the top-left corner of the two images until the bottom-right corner is reached

The overall saliency of a window is defined as
$$C(w) = \max(s(a|w), s(b|w)) \qquad 30$$

$$c(w) = \frac{C(w)}{\left(\sum_{w' \in /w} C(w')\right)} \qquad 31$$

The Keywords section begins with the word, "Keywords" in 13 pt. Times New Roman, bold italics, "Small Caps" font with a 6pt. spacing following. There may be up to five keywords (or short phrases) separated by commas and six spaces, in 10 pt. Times New Roman italics. An 18 pt. line spacing follows.

## 6. RESULT ANALYSIS

Image Fusion techniques requires the registered images for testing. Image Registration [4] is the determination of a geometrical transformation that aligns points in one view of an object with corresponding points in another view of that object or another object. The experiments are carried out with the registered images.

The various statistical rules have been analyzed and the proposed statistical fusion rule (WAMM) is tested in both wavelet domain and NSCT domain on Medical images, Remote Sensing images and Multi Focus images.

## 6.1. Experiment 1 on Medical Images

Different medical imaging techniques may provide scans with complementary and occasionally conflicting information. The combination of images can often lead to additional clinical information not apparent in the separate images. The goal of image fusion is to impose a structural anatomical framework in functional images. Often a single functional image may not contain enough anatomical details to determine the position of a tumour or other lesion.

### 6.1.1. Aim:

To fuse two greyscale medical images, of which one is a CT image and the other is a MR image.

### 6.1.2. Experimental Setup:

*Input images:* 256 x 256 greyscale CT and MR image of brain (Figure 5.1 (a-b)).

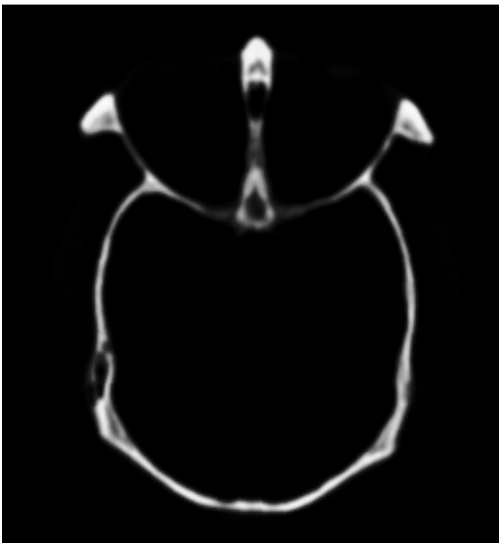 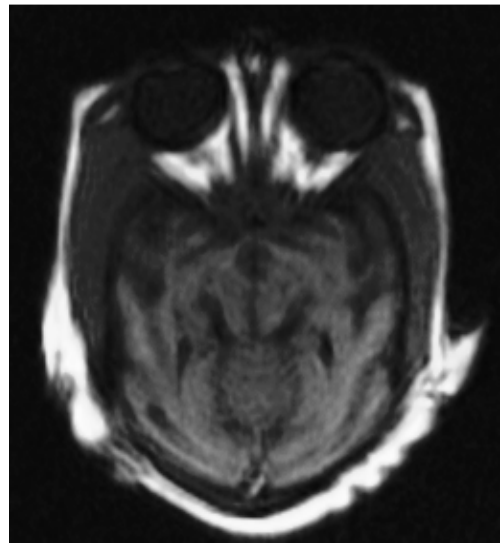

(a)   (b)

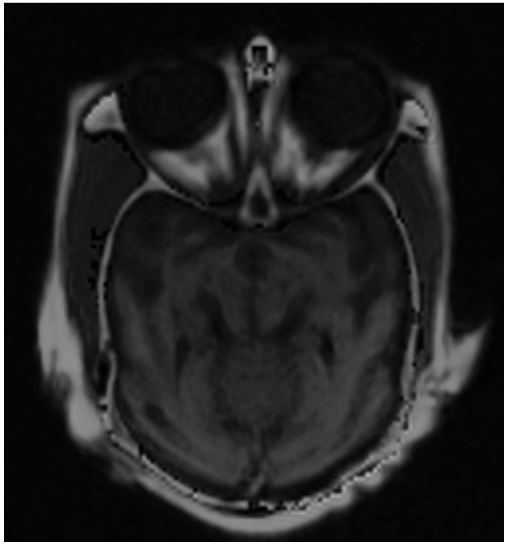 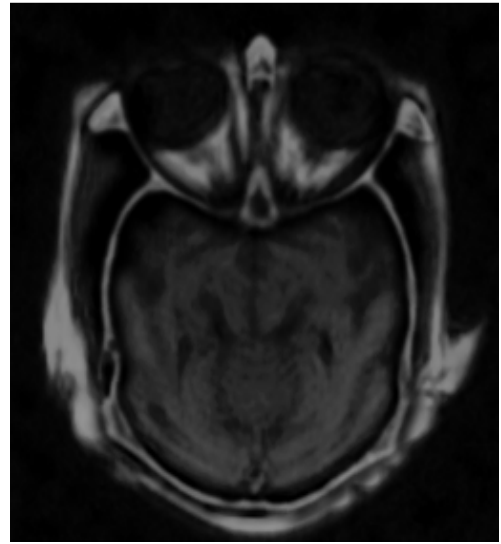

(c) (d)

Figure 5:(a)CT image (b)MR image (c)Wavelet fused image using Entropy

(d) NSCT fused image using Entropy

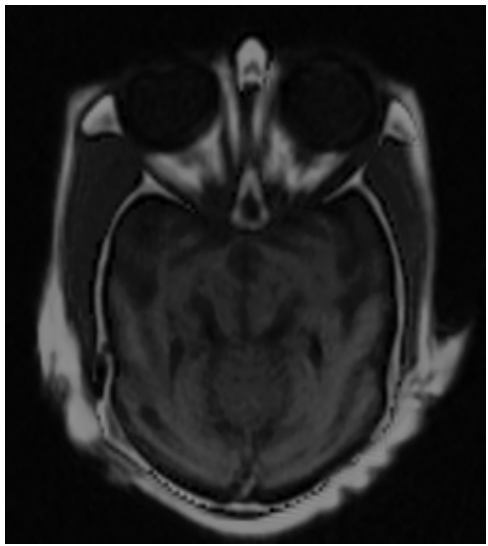 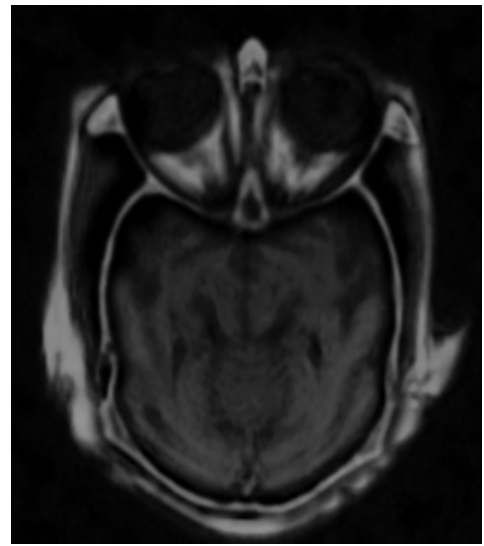

(e) (f)

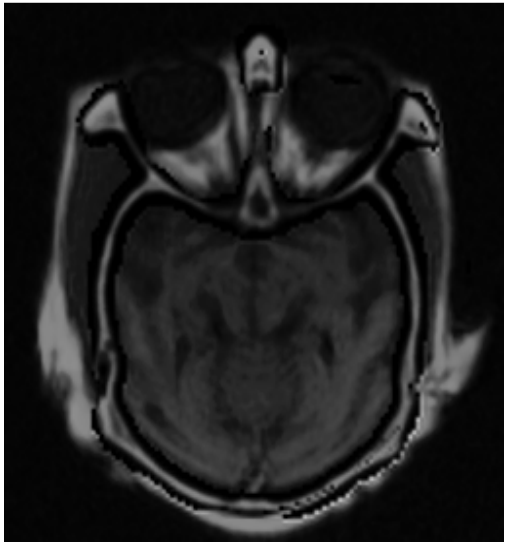 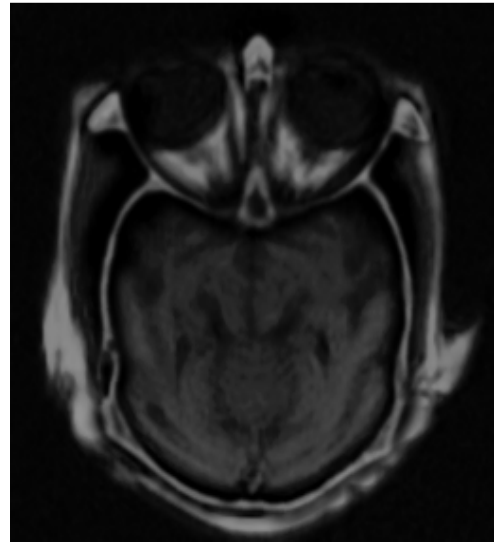

(g) (h)

Figure 5:(e) Wavelet fused image using Mean(f) NSCT fused image using Mean

(g) Wavelet fused image using SD   (h) NSCT fused image using SD

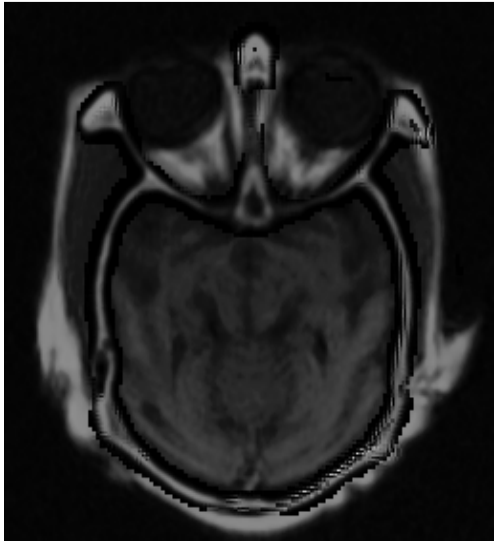 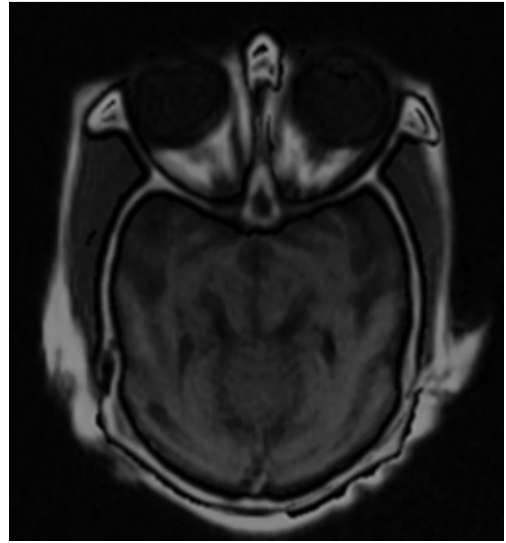

(i) (j)

Figure 5:(i) Wavelet fused image using WAMM (j)NSCT fused image using WAMM

### 6.1.3 Comparative Analysis

Table 1 : The performance measures obtained for Medical Image Fusion using different methods

|  | Domain | EN1 | EN2 | EN3 | S | PM |
| --- | --- | --- | --- | --- | --- | --- |
| Method1 (Entropy) | Wavelet | 1.7126 | 5.6561 | 2.4847 | 0.4795 | 0.5887 |
|  | NSCT | 1.7126 | 5.6561 | 6.0831 | 0.5630 | 0.6117 |
| Method2 (Mean) | Wavelet | 1.7126 | 5.6561 | 5.8754 | 0.5497 | 0.6782 |
|  | NSCT | 1.7126 | 5.6561 | 5.9090 | 0.5743 | 0.7547 |
| Method3 (S D) | Wavelet | 1.7126 | 5.6561 | 5.8615 | 0.4877 | 0.6174 |
|  | NSCT | 1.7126 | 5.6561 | 5.9090 | 0.5618 | 0.6383 |
| WAMM | Wavelet | 1.7126 | 5.6561 | 5.8595 | 0.4882 | 0.6225 |
|  | NSCT | 1.7126 | 5.6561 | 5.9090 | 0.5625 | 0.6373 |

Here EN1 and EN2 represent the entropy of the original images to be fused in wavelet and NSCT domain respectively. In the fused image with, WAMM performs better than in NSCT than wavelet domain and it preserves more details in the fused image. The artifacts and inconsistencies in wavelet domain is removed in NSCT domain using WAMM method. In the above table, it is seen that the fusion with SD, Similarity and PM gives better results

### 6.2. Experiment 2 on Multifocus images

Due to the limited depth-of-focus of optical lenses (especially those with long focal lengths) it is often not possible to get an image that contains all relevant objects "in focus". One possibility to overcome this problem is to take several pictures with different focus points and combine them together into a single frame that finally contains the focused regions of all input images.

### 6.2.1. Aim:

To fuse two greyscale multifocus images using the existing methods and the proposed method.

### 6.2.2. Experimental Setup:

*Input images:* 256 x 256 greyscale clock images with Figure 5.2:(a)Focus on right clock (b)Focus on left clock.

**6.2.3. Results:**

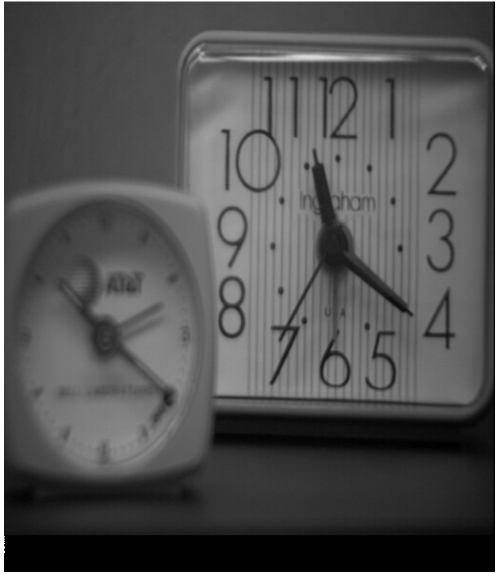 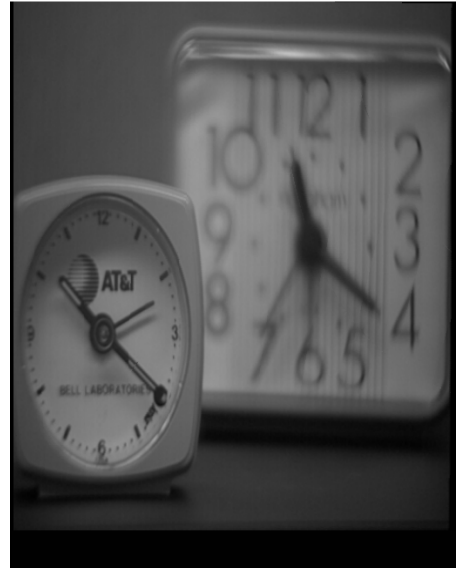

(a) (b)

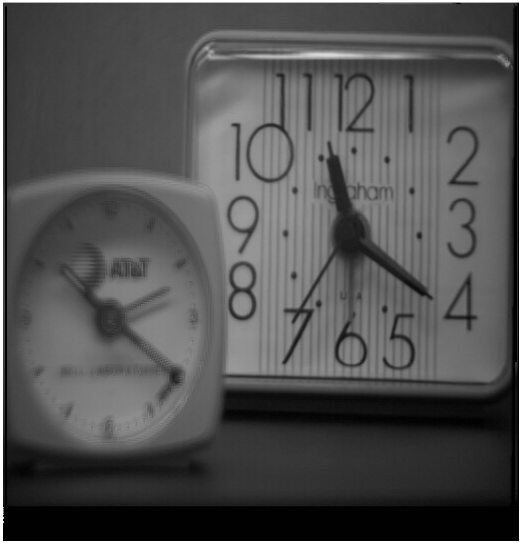 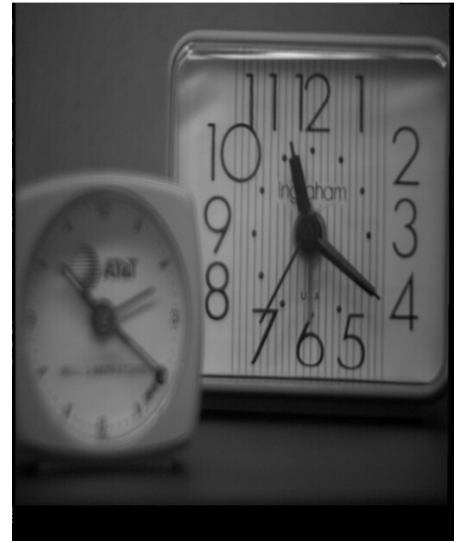

(c) (d)

Figure 6:(a)Focus on right clock (b)Focus on left clock (c)Wavelet fused image using Entropy
(d) NSCT fused image using Entropy

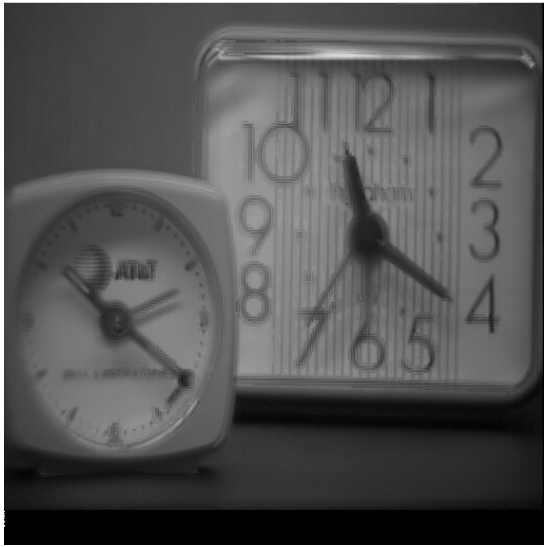 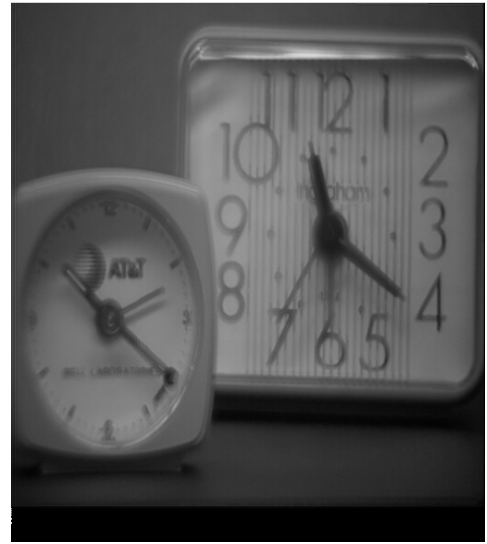

(e)            (f)

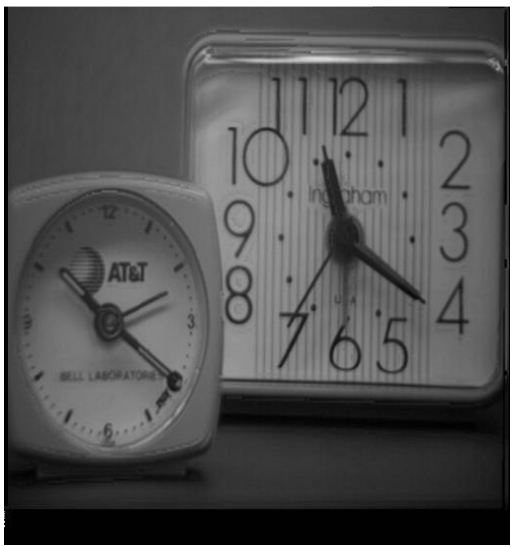 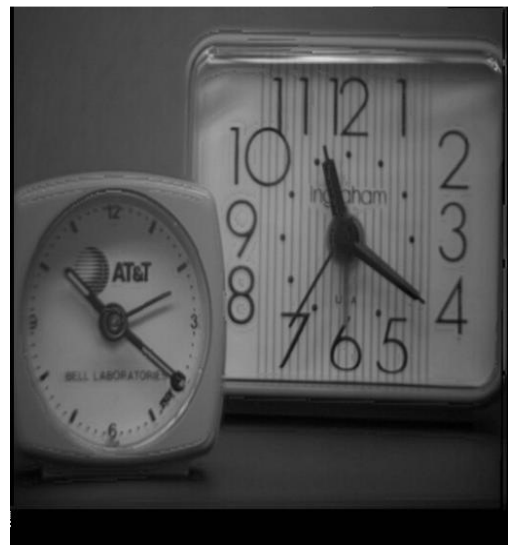

(g)            (h)

Fig 6 (e) Wavelet fused image using Mean(f) NSCT fused image using Mean(g) Wavelet fused image using SD(h) NSCT fused image using SD

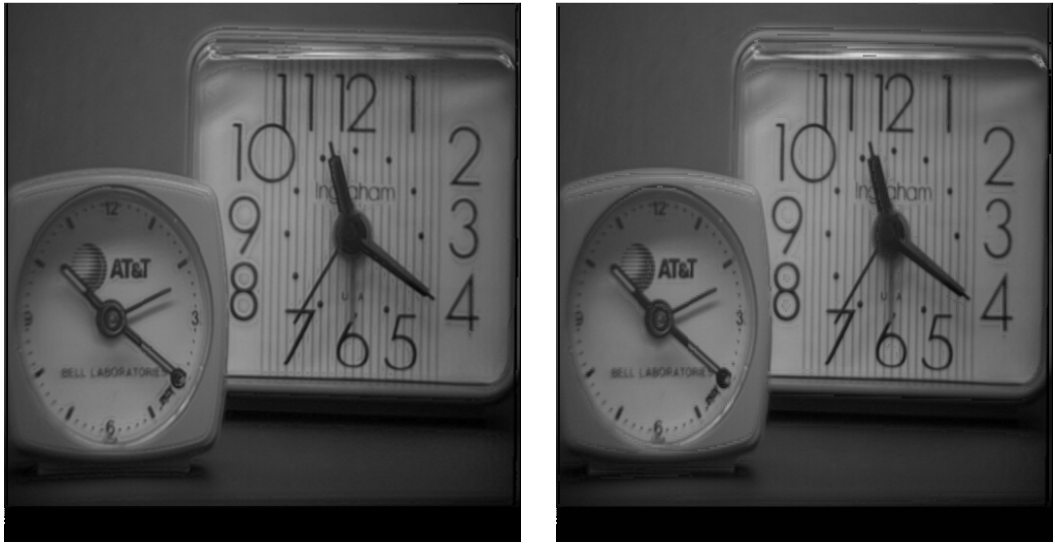

(i)                                               (j)

Fig 6 (i) Wavelet fused image using WAMM (j) NSCT fused image using WAMM

**6.2.3. Comparative Analysis:**

Table 2 : The performance measures obtained for multi-focus image fusion using different methods

|  | Domain | EN1 | EN2 | EN3 | S | PM |
|---|---|---|---|---|---|---|
| **Method1** (Entropy) | Wavelet | 6.9803 | 6.9242 | 7.0052 | 0.6462 | 0.5317 |
| | NSCT | 6.9803 | 6.9242 | 7.0260 | 0.6317 | 0.5623 |
| **Method2** (Mean) | Wavelet | 6.9803 | 6.9242 | 6.9925 | 0.5039 | 0.4286 |
| | NSCT | 6.9803 | 6.9242 | 7.0429 | 0.6111 | 0.5258 |
| **Method3** (SD) | Wavelet | 6.9803 | 6.9242 | 6.9960 | 0.5983 | 0.5508 |
| | NSCT | 6.9803 | 6.9242 | 7.0386 | 0.6349 | 0.5614 |
| **WAMM** | Wavelet | 6.9803 | 6.9242 | 6.9980 | 0.5970 | 0.5513 |
| | NSCT | 6.9803 | 6.9242 | 7.0421 | 0.6379 | 0.5640 |

Here EN1 and EN2 represent the entropy of the original images to be fused in wavelet and NSCT domain respectively

In the fused image with, WAMM performs better than in NSCT than wavelet domain and it preserves more details in the fused image. The artifacts and inconsistencies in wavelet domain is removed in NSCT domain using WAMM method. In the above table, it is seen that the fusion with SD, Similarity and PM gives better results.

## 7. CONCLUSIONS

In this work, a new statistical fusion rule, Weighted Average Merging Method (WAMM) is proposed in NSCT domain. A review of the different statistical fusion rules such as Entropy , Mean ,Standard Deviation and Weighted Arithmetic Merging Method is discussed. In the method, fusion rule using Standard Deviation, the edge information is preserved successfully within the image, but image lacked visual quality that we expected. So in order to overcome the limitations of the existing statistical fusion rules, the combination of Standard Deviation in the coarse scales and Weighted Arithmetic Merging Method (WAMM) in the fine scales is proposed.

A new statistical fusion rule, WAMM is proposed in NSCT domain. Experimental results shows that WAMM method for image fusion obtained better results in NSCT domain when tested with performance measures, SD, Similarity and Piella Metric. It preserves the edge details and the visual quality of the fused image. The analysis obtained shows that the proposed WAMM yields better results in NSCT domain. This proposed scheme is tested both in NSCT and in wavelet domain and the results were compared and obtained better results. As a future work, an Adaptive Weighted Average Merging Method can be suggested.

## ACKNOWLEDGEMENTS

All authors would like to thank Dr. Oliver Rockinger and the TNO Human Factors Research Institute for providing the source IR and visible images that are publicly available online at http://www.imagefusion.org.

The International Journal of Multimedia & Its Applications (IJMA) Vol.4, No.2, April 2012

**Authors**

Manu.V.T. was born in Kerala, India on June 19, 1985. He graduated in Computer Science & Engineering from National Institute of Technology, Calicut, India in 2007. He did his masters at University of Kerala with specialization in Digital Image Computing. His area of interest are image processing and operating systems .

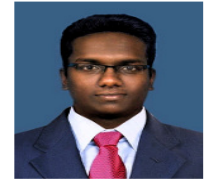

PhilominaSimon did her B.Tech(Govt Engg College, Thiruvananthapuram), and M.Tech CSE from Pondicherry University. Her area of interest are artificial intelligence,computer networks and image processing

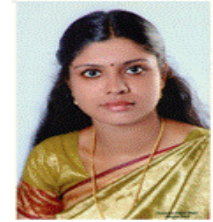